# OSTA: One-shot Task-adaptive Channel Selection for Semantic Segmentation of Multichannel Images

Yuanzhi Cai, *Member, IEEE,* Jagannath Aryal, *Member, IEEE,* Yuan Fang, Hong Huang, and Lei Fan, *Member, IEEE*

*Abstract*—Semantic segmentation of multichannel images is a fundamental task for many applications. Selecting an appropriate channel combination from the original multichannel image can improve the accuracy of semantic segmentation and reduce the cost of data storage, processing and future acquisition. Existing channel selection methods typically use a reasonable selection procedure to determine a desirable channel combination, and then train a semantic segmentation network using that combination. In this study, the concept of pruning from a supernet is used for the first time to integrate the selection of channel combination and the training of a semantic segmentation network. Based on this concept, a One-Shot Task-Adaptive (OSTA) channel selection method is proposed for the semantic segmentation of multichannel images. OSTA has three stages, namely the supernet training stage, the pruning stage and the fine-tuning stage. The outcomes of six groups of experiments (L7Irish3C, L7Irish2C, L8Biome3C, L8Biome2C, RIT-18 and Semantic3D) demonstrated the effectiveness and efficiency of OSTA. OSTA achieved the highest segmentation accuracies in all tests (62.49% (mIoU), 75.40% (mIoU), 68.38% (mIoU), 87.63% (mIoU), 66.53% (mA) and 70.86% (mIoU), respectively). It even exceeded the highest accuracies of exhaustive tests (61.54% (mIoU), 74.91% (mIoU), 67.94% (mIoU), 87.32% (mIoU), 65.32% (mA) and 70.27% (mIoU), respectively), where all possible channel combinations were tested. All of this can be accomplished within a predictable and relatively efficient timeframe, ranging from 101.71% to 298.1% times the time required to train the segmentation network alone. In addition, there were interesting findings that were deemed valuable for several fields.

*Index Terms*— band selection, channel selection, cloud detection, domain adaptation, Landsat, multichannel images, multispectral images, semantic segmentation, supervised band selection.

This work was supported in part by the Xi'an Jiaotong-Liverpool University Research Enhancement Fund under Grant REF-21-01-003, in part by the Xi'an Jiaotong-Liverpool University Key Program Special Fund under Grant KSF-E-40, and in part by the Xi'an Jiaotong-Liverpool University Research Development Fund under Grant RDF-18-01-40. *(Corresponding author: Lei Fan.)*

Yuanzhi Cai, Yuan Fang and Hong Huang are with the Department of Civil Engineering, Design School, Xi'an Jiaotong-Liverpool University, Suzhou 215000, China, and also with the School of Engineering, University of Liverpool, L69 3BX Liverpool, U.K. (e-mail: yuanzhi.cai19@student.xjtlu.edu.cn; yuan.fang16@student.xjtlu.edu.cn; hong.huang19@student.xjtlu.edu.cn).

Jagannath Aryal is with the Department of Infrastructure Engineering, Faculty of Engineering and IT, The University of Melbourne, Melbourne VIC 3010, Australia. (e-mail: jagannath.aryal@unimelb.edu.au).

Lei Fan is with the Department of Civil Engineering, Design School, Xi'an Jiaotong-Liverpool University, Suzhou 215000, China. (e-mail: lei.fan@xjtlu.edu.cn).

## I. INTRODUCTION

**M**ULTICHANNEL image refers to image data containing more than three channels (typically less than ten channels). It is also known as multispectral image when only spectral channels are included. Semantic segmentation of multichannel image is the basis for many remote sensing applications, such as cloud detection [1]–[3], land use/ land cover classification [4], [5] and forest monitoring [6], [7].

Although additional channels provide more information for semantic segmentation, it is not desirable to use extra channels than are necessary for many practical applications. Firstly, there is always a cost associated with the acquisition of additional channels. For example, the quest for additional spectral channel(s) by multispectral sensors not only entails higher financial costs, but often leads to compromises in spatial resolutions. Secondly, higher segmentation accuracy can be achieved by excluding unnecessary channels that are impacted by noise and false information, and excessive less relevant channels [8]–[14].

For some widely recognised classes of objects (e.g., clouds and water bodies), their preferred channel(s) for segmentation have been studied for decades. However, these studies are insufficient to meet the challenges in demand for semantic segmentation. On one hand, it is often required to segment multiple classes simultaneously. On the other hand, more refined object classes and new data are constantly introduced for segmentation. Therefore, an automated channel selection method that can adapt to the needs of different tasks is highly desirable.

Currently, the main type of imagery data investigated in the field of channel selection is hyperspectral images. In this context, a channel selection task often needs to select dozen(s) of channels out of one to two hundred candidate channels. Due to the fine spectral resolution of a hyperspectral image, its neighbouring channels often contain features that are very similar to each other. Consequently, it is possible to achieve semantic segmentation accuracy in the satisfactory range using hyperspectral images with removal of redundant channels and can sometimes achieve even higher accuracy than using the original hyperspectral channels [15]–[19]. Coupled with the fact that the removal of redundant channels does not rely on labelling information, unsupervised methods have become the dominant research direction for channel selection. In unsupervised methods, channel selection is typically performed by using one or a combination of ranking, clustering and search strategies to optimise various criteria, such as entropy [20]–[22], variants of PCA [23]–[25], and minimising similarity [26]–[29]. These methods select channels based on the characteristics of the input data themselves and do not take into account the preferences of classes to be segmented. In other words, they are not task-adaptive.

There are also supervised channel selection methods that make use of label information. The major difference between supervised and unsupervised methods lies in their optimisation criteria. More specifically, additional criteria such as prediction accuracy [30], mutual information criteria [31]–[34] and Fisher score [35] became available to supervised methods due to the presence of label information. Although existing supervised methods are task-adaptive, they often suffer from the following common deficiencies. Firstly, most methods use criteria other than prediction accuracy, which diverts the optimisation process from obtaining the highest prediction accuracy and is likely to lead to selecting channel combinations with sub-optimal accuracy. Secondly, a majority (sometimes all) of excluded channels in most channel selection strategies were determined by evaluating individual channels. Since "the $m$ best features are not the best $m$ features" [34], [36], [37], using such selection strategies is not the best solution. Finally, most of the existing supervised methods require training multiple classifiers and/or training classifier(s) multiple times, which leads to low computational efficiency.

As an attempt to resolve the aforementioned issues associated with existing supervised methods, a novel one-shot task-adaptive (OSTA) channel selection method is proposed in this study. OSTA has the following characteristics: a) directly optimising for segmentation accuracy, b) considering channel interactions (i.e., no channel is excluded individually), c) integrating channel selection and network fine-tuning within a relatively efficient and predicable timeframe. All these characteristics are realised by formulating the channel selection as a pruning process for a supernet. As such OSTA consists three stages, namely the training stage of the supernet, the pruning stage, and the fine-tuning stage. In this study, the effectiveness and the efficiency of OSTA are tested using four datasets, including an eight-band cloud detection dataset named L7 Irish [38], a ten-band cloud detection dataset named L8 Biome [39], a six-band very-high resolution dataset named RIT-18 [40] and an eight-channel image dataset that is transformed from a terrestrial laser scanning (TLS) point cloud dataset named Semantic3D [41]. The major contributions of this paper are:

1)  Development of a novel channel selection method (i.e., OSTA) to overcome the following main issues with the existing methods. They are not dedicated to optimising segmentation accuracy, do not take full account of channel interactions and require repeated training of classifier(s).

2)  Comprehensive evaluation of channel selection methods by exhaustive testing of the accuracy performance of 3-channel combinations on four benchmark datasets.

3)  In addition to channel selection, this study also leads to some interesting findings: a) there are channel combinations that are robust to the network initialisation; b) the coastal aerosol band has been neglected in the past research for cloud detection but turns out to be an importance channel for cloud detection according to our study; c) training with channel combinations other than the selected one can improve the semantic segmentation accuracy.

## II. METHODOLOGY

### A. OSTA

The overall training strategy of OSTA is shown in Fig.1. It starts with a supernet training stage that accounts for 15% of the total training iterations. At this first stage, the objective of the supernet training is to perform semantic segmentation using any combination of channels, and during the training the learning rate increases linearly from zero to the target value. Subsequently, the channel combinations used for training are progressively pruned according to their validation accuracies until only one combination remains. This pruning process accounts for 35% of the total training iterations. In the final stage, the remaining 50% of the total training iterations are used to fine-tune the semantic segmentation network (SSN) for the selected channel combination. The poly learning rate policy is used in the latter two stages.

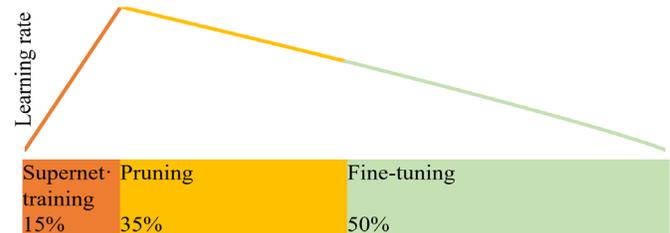

**Fig. 1.** Percentages of training iterations of the three stages in OSTA and the learning rate schedule.

The essence of OSTA is to select the channel combination with the best validation accuracy. This can ensure two points: a) a good generalisation performance (i.e., accuracy) of the selected channel combination, b) the combination of "best $m$ features" is taken into account as none of the channels is removed individually. More detailed explanations of these three stages are as follows.

#### 1)  Supernet training

The supernet has two parts: an input layer that is pruned later, which treats each channel combination as one of its input channels (ICs), and a subsequent weight-sharing SSN which takes individual inputs from each input channel (IC) for semantic segmentation. The key to a successful supernet training is to train each candidate channel (also known as path or branch) fairly, to which many studies have been devoted [42]–[46]. Fortunately, fair training of candidate channels can be easily achieved in OSTA by randomly sampling ICs during the training. Therefore, pruning from a supernet is inherently suitable for implementing channel selection.

TABLE I
SUMMARY OF BENCHMARK DATA USED

| Dataset | No. classes | No. channels | No. and size of image(s) | | Crop size | No. cropped images | | |
|---|---|---|---|---|---|---|---|---|
| | | | Training | Testing | | Sub-training | Sub-validation | Testing |
| L7 Irish 2C | 2 | 8 | 155, ≈8000×8000 | 51, ≈8000×8000 | 1024×1024 | 10012 | 100 | 3246 |
| L7 Irish 3C | 3 | 8 | 155, ≈8000×8000 | 51, ≈8000×8000 | 1024×1024 | 10012 | 100 | 3246 |
| L8 Biome 2C | 2 | 10 | 72, ≈8000×8000 | 24, ≈8000×8000 | 1024×1024 | 5041 | 50 | 1555 |
| L8 Biome 3C | 3 | 10 | 72, ≈8000×8000 | 24, ≈8000×8000 | 1024×1024 | 5041 | 50 | 1555 |
| RIT-18 | 18 | 6 | 1, 9393×5642 | 1, 8833×6918 | 1024×1024 | 48 | 12 | 63 |
| Semantic3D | 8 | 8 | 10, 3600×7200 | 5, 3600×7200 | 1024×2048 | 128 | 32 | 80 |

### 2) Pruning

To minimise the additional amount of computation required for pruning, a discrete forward-only pruning strategy that is based on validation accuracy is proposed. More specifically, the training of the SSN is paused uniformly for $n$ times during the pruning stage, where $n$ is the number of ICs to be pruned. At each pause, the segmentation accuracy of the SSN on the validation set is tested for each remaining IC. The IC with the worst validation accuracy is removed in subsequent training. By repeating the training of the SSN and the pruning of an IC in sequence, the selected channel combination (SCC) (i.e., selected IC) is obtained at the end of the pruning stage. Since only one layer in the supernet needs to be pruned, the validation accuracy can serve as the equivalent of gradients in OSTA.

### 3) Fine-tuning

This stage is to fine-tune the SSN on the SCC. Since validation accuracy is not required for this stage, the validation set can be merged into the training set to fine-tune the SSN.

### B. Establishment of benchmarks

#### 1) Benchmark data

Four semantic segmentation datasets are used in this study, including L7 Irish [38], L8 Biome [39], RIT-18 [40] and Semantic3D [41], and they are used as follows.

L7 Irish and L8 Biome are processed in the same way as both datasets contain multispectral satellite images and are labelled with the same 4 classes (i.e., shadow, clear sky, thin cloud and thick cloud). Each dataset is partitioned evenly into a training set (75%) and a test set (25%). More specifically, the last of every four images is used for testing, according to the order of the images on their official website. As for the label configurations, two commonly used ones [3], [47] are tested in this study. In the first configuration, classes of shadow and clear sky are merged into a single class of background. Datasets with this configuration (i.e., background, thin cloud and thick cloud) are denoted as L7 Irish 3C and L8 Biome 3C. The second configuration is established upon the first one, where classes of thin cloud and thick cloud are further merged into a single class of cloud. Datasets with this configuration (i.e., background and cloud) are denoted as L7 Irish 2C and L8 Biome 2C. For L7 Irish and L8 Biome, the bands used in this study are blue (B), green (G), red (R), near-infrared (NIR), short-wave infrared (SWIR), thermal low gain (TLG), thermal high gain (THG) and mid-infrared (MIR), and coastal aerosol (CA), B, G, R, NIR, SWIR1, SWIR2, cirrus (C), thermal 1 (T1) and thermal 2 (T2), respectively.

RIT-18 is a six-band (B, G, R, NIR1, NIR2, NIR3) image dataset having 18 labelled classes. The original validation set of RIT-18 is used as the test set in this study.

Semantic3D is originally a point cloud dataset having 8 labelled classes, which is transformed into an eight-channel image dataset in this study. The original training set of Semantic3D is partitioned into a training set and a test set as in the previous study [8]. Each point cloud data is transformed into an eight-channel image using spherical projection and the enhancement method proposed in [9]. Specifically, the following feature channels are included: R, G, B, intensity (I), z-coordinate image (Z), depth image (D), enhanced z-coordinate image (Ze), enhanced depth image (De).

The key characteristics of benchmark data used are summarised in Table I. Since the original images are too large to be fed into the network, they are cropped into smaller images without overlaps. The crop sizes are shown in Table I. To calculate the validation accuracy, the training set is partitioned into a sub-training set and a sub-validation set with similar class distribution in the first two stages of OSTA.

#### 2) Benchmark methods

There are three types of benchmark methods used in this study. They are briefly introduced in this section.

The first one is the direct feeding (DF), which uses the original multichannel image as the input data for the SSN. DF is considered as a baseline method.

The second type of methods are channel selection methods, which select the "best channel combination" as input data for the SSN. Different channel selection methods will result in different "best channel combination" depending on the selection criteria and selection strategy used. To provide the most comprehensive comparison, we exhaustively tested all possible channel combinations using supervised grid search (SGS). Although the SGS test results have implicitly included all the segmentation accuracies that can be achieved using existing channel selection methods, we also explicitly compared three recent channel selection methods, including BS-Nets [12], ONR [13] and DARecNet-BS [48].

The third type of methods are feature extraction methods,

TABLE II
EVALUATION METRICS USED

| | | |
|---|---|---|
| Accuracy | Mean accuracy (mA) | $\dfrac{1}{C}\sum_{c=1}^{C}\dfrac{TP_c}{\text{Number of points}_c}$ |
| | Mean intersection over union (mIoU) | $\dfrac{1}{C}\sum_{c=1}^{C}\dfrac{TP_c}{TP_c + FP_c + FN_c}$ |
| | Combination accuracy percentile (CAP) | Percentage of SGS-derived combinations with test accuracies lower than OSTA accuracy. For an OSTA accuracy that is not identical to the SGS accuracy, its CAP is obtained by linear interpolation based on the two nearest larger and smaller SGS accuracies. |
| | Difference in combination accuracy (DAC) | Accuracy of $SCC_{OSTA}$ − Accuracy of $SCC_{OSTA}$ in SGS |
| Efficiency | Ratio of additional time (RAT) | $\dfrac{\text{Time of training OSTA}}{\text{Time of direct training SSN}} - 100\%$ |
| | Ratio of additional memory (RAM) | $\dfrac{\text{Memory of training OSTA}}{\text{Memory of direct training SSN}} - 100\%$ |

Where $TP_c$, $FP_c$, and $FN_c$ represent the true positive, false positive and false negatives of class c, respectively.

which use the extracted new feature channels as input data for the SSN. Although feature extraction methods have been criticized for not preserving the original channels, it is still interesting to test the segmentation accuracy they can obtain. Two feature extraction methods are used for benchmarking in this study, namely principal component analysis (PCA) and LC-Net [49]. The former is a classical feature extraction method, while the latter is a task-adaptive feature extraction method that performs a linear compression of the input data dimensions by learning.

3) **Training setting**

For a fair comparison, the same SSN is used for all tested methods. Specifically, the backbone and decoder of SSN are ConvNeXt-T [50] and UperNet [51], respectively, given their proven performance [52]. Unless otherwise specified, ConvNeXt-T is initialised with the ImageNet pre-trained weights. The training strategy used is the same as the original implementation [50] except for the following points. The total number of training iterations is set as 10k. The sizes of the input patches used for the network training are set to 1/4 of the crop size in Table I, i.e., 512×512 for L7 Irish, L8 Biome and RIT-18, and 512×1024 for Semantic3D. It is worth noting that the segmentation accuracy can significantly be improved by adjusting the histogram of the benchmark datasets to match the distribution of the pre-training dataset, which is implemented in this study. Finally, the accuracy evaluation for the Semantic3D data is based on the accuracy of the image results.

Except for DF, both channel selection and feature extraction methods require a predetermined dimension reduction target. In this study, the dimension reduction target of three was tested, i.e., the number of channels of the original multichannel image was reduced to three. The rationale for testing this setup is as follows. Firstly, for two of the benchmark data used (i.e., RIT-18 and Semantic3D), previous studies [9], [49] showed that a better segmentation accuracy was achieved using a properly selected three channels than that using more channels. Secondly, higher accuracy can be obtained when fine-tuning with data similar to the pre-training data. Coupled with the fact that most state-of-the-art computer vision models are pre-trained on RGB datasets, it is often desirable to reduce the dimension of multichannel images to 3 in practical applications. For example, compressing four-channel images (i.e., red, green, blue and thermal) into three-channel images (i.e., 0.5×(red + thermal), 0.5×(green + thermal), 0.5×(blue + thermal)) has become a popular fusion method in the field of crack detection [53]–[55].

### C. Evaluation metrics

In total, six evaluation metrics are used in this study to measure the performance of OSTA, as shown in Table II. Mean accuracy (mA) and mean intersection over union (mIoU) are used as the major accuracy metrics for RIT-18 and other three datasets (i.e., L7 Irish, L8 Biome, and Semantic3D), respectively. In addition, two novel accuracy metrics are proposed in this study, namely combination accuracy percentile (CAP) and difference in combination accuracy (DCA). CAP measures the ranking of OSTA test accuracy against the accuracies of all combinations tested in SGS. This metric represents the percentage of SGS combinations that their test accuracies are exceeded by OSTA. DCA measures the difference between the test accuracy of "SCC in OSTA" ($SCC_{OSTA}$) and the test accuracy of $SCC_{OSTA}$ in SGS. This metric describes the impact of the first two training stages of OSTA on the test accuracy of $SCC_{OSTA}$. This indicator can help better understand the role of the different stages of OSTA.

The measure of efficiency is based on the ratio of consumptions (time and memory) with OSTA to consumptions without OSTA (i.e., direct training SSN). This allows the evaluation metrics to show the additional consumption caused by channel selection in percentage.

### III. Experiments and results

#### A. Semantic segmentation on benchmarks

The qualitative semantic segmentation results are visualised in Fig.2, where the horizontal and vertical coordinates show the relative accuracy (CAP) and absolute accuracy (mIoU/mA), respectively. Each data point regarding the SGS method represents the segmentation accuracy of each three-channel combination tested.

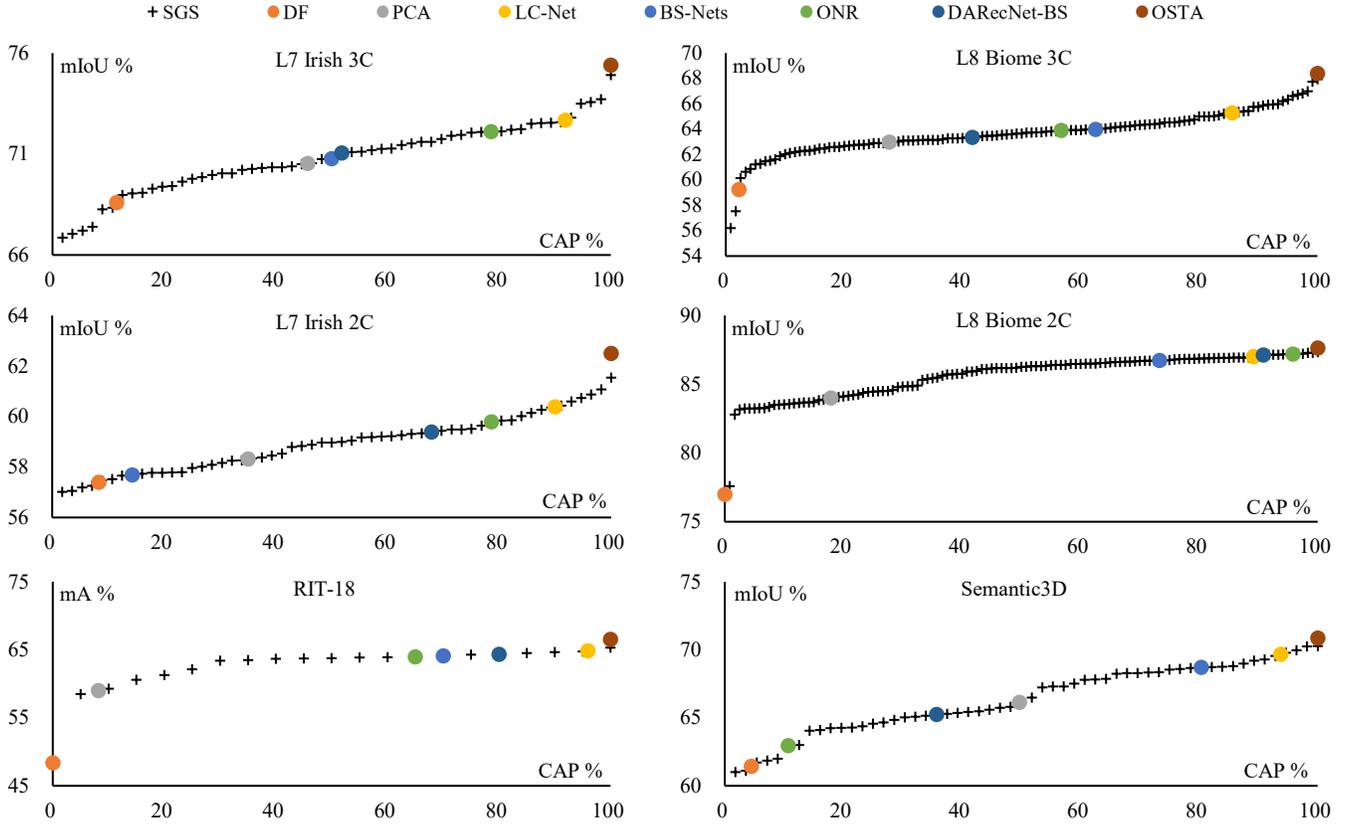

**Fig. 2.** Semantic segmentation results on benchmark data.

TABLE III
SUMMARY OF THE SEMANTIC SEGMENTATION RESULTS ON BENCHMARK DATA

| | | L7 Irish 3C | | | L8 Biome 3C | | | RIT-18 | | |
|---|---|---|---|---|---|---|---|---|---|---|
| | | Index | mIoU % | CAP % | DCA % | Index | mIoU % | CAP % | DCA % | Index | mA % | CAP % | DCA % |
| DF | | - | 57.39 | 8.33 | - | - | 59.24 | 2.21 | - | - | 48.41 | 0 | - |
| Feature extraction | PCA | - | 58.31 | 35.04 | - | - | 62.96 | 27.64 | - | - | 58.99 | 8.21 | - |
| | LC-Net | - | 60.38 | 89.96 | - | - | 65.27 | 85.56 | - | - | 64.86 | 95.96 | - |
| Channel selection | BS-Nets | 42 | 57.68 | 14.29 | 0 | 30 | 63.97 | 62.50 | 0 | 4 | 64.09 | 70.00 | 0 |
| | ONR | 36 | 59.78 | 78.57 | 0 | 81 | 63.89 | 56.67 | 0 | 7 | 63.99 | 65.00 | 0 |
| | DARecNet-BS | 32 | 59.38 | 67.86 | 0 | 82 | 63.35 | 41.67 | 0 | 3 | 64.33 | 80.00 | 0 |
| | SGS | 50 | 61.54 | 100.00 | 0 | 3 | 67.94 | 100.00 | 0 | 19 | 65.32 | 100.00 | 0 |
| | OSTA | 56 | 62.49 | 100.00 | +1.42 | 9 | 68.38 | 100.00 | +1.69 | 19 | 66.53 | 100.00 | +1.21 |

| | | L7 Irish 2C | | | L8 Biome 2C | | | Semantic3D | | |
|---|---|---|---|---|---|---|---|---|---|---|
| | | Index | mIoU % | CAP % | DCA % | Index | mIoU % | CAP % | DCA % | Index | mIoU % | CAP % | DCA % |
| DF | | - | 68.59 | 11.45 | - | - | 76.97 | 0 | - | - | 61.44 | 4.47 | - |
| Feature extraction | PCA | - | 70.52 | 45.71 | - | - | 83.99 | 17.92 | - | - | 66.13 | 49.68 | - |
| | LC-Net | - | 72.67 | 91.84 | - | - | 87.00 | 89.17 | - | - | 69.67 | 93.79 | - |
| Channel selection | BS-Nets | 42 | 70.76 | 50.00 | 0 | 30 | 86.72 | 73.33 | 0 | 34 | 68.72 | 80.36 | 0 |
| | ONR | 36 | 72.10 | 78.57 | 0 | 81 | 87.19 | 95.83 | 0 | 27 | 62.96 | 10.71 | 0 |
| | DARecNet-BS | 32 | 71.04 | 51.79 | 0 | 82 | 87.12 | 90.83 | 0 | 43 | 65.26 | 35.71 | 0 |
| | SGS | 56 | 74.91 | 100.00 | 0 | 106 | 87.32 | 100.00 | 0 | 56 | 70.27 | 100.00 | 0 |
| | OSTA | 50 | 75.40 | 100.00 | +1.69 | 23 | 87.63 | 100.00 | +0.42 | 36 | 70.86 | 100.00 | +1.08 |

DF is the worst one among all benchmark methods. For all benchmark data, a randomly selected 3-channel combination would likely achieve higher accuracy than DF. This demonstrates the value of band selection for semantic segmentation of multichannel images. The reason for this is that fine-tuning with data that has the same channel number as the pre-training data can result in higher accuracy, as mentioned in Section II.B.3.

It was noticed that the CAPs of existing dimension reduction methods (i.e., PCA, LC-Net, BS-Net, ONR and DARecNet) were unstable across benchmark data. This indicates that they are short of task adaptability. In contrast, the proposed OSTA achieved the highest accuracy (i.e., saturated CAP of 100%) in all tests, which proved its effectiveness. It was surprising that even OSTA was able to outperform highest SGS accuracy. Since existing channel selection methods train the SSN with the SCC alone, the highest SGS accuracy is usually considered to be the upper limit achievable. Investigating the reason why OSTA can exceed this upper limit is one of the focuses of subsequent ablation study (Section III.C).

TABLE IV
MEASURED EFFICIENCY METRICS AND CALCULATION FOR ESTIMATING RAT

| | | L7 Irish | L8 Biome | RIT-18 | Semantic3D |
|---|---|---|---|---|---|
| RAM | | 0 | 0 | 0 | 0 |
| RAT | Calculation steps for estimation | **Patches to be processed in each epoch of sub-validation set** $100 \times 4 = 400$ | $50 \times 4 = 200$ | $12 \times 4 = 48$ | $32 \times 4 = 128$ |
| | | **Total number of 3-channel combinations** 56 | 120 | 20 | 56 |
| | | **Total number of epochs of sub-validation set during pruning** $56 + 55 + \ldots + 2 = 1595$ | $120 + 119 + \ldots + 2 = 7259$ | $20 + 19 + \ldots + 2 = 209$ | $56 + 55 + \ldots + 2 = 1595$ |
| | | **Total equivalent pruning iterations** $400 \times 1595 / 16 = 39875$ | $200 \times 7259 / 16 = 90737.5$ | $48 \times 209 / 16 = 627$ | $128 \times 1595 / 16 = 12,760$ |
| | | **Ratio between equivalent pruning and training iterations** 398.8% | 907.4% | 6.27% | 127.6% |
| | | **Ratio between time of a pruning and a training iteration (measured)** 20.9% | 21.7% | 21.5% | 25.5% |
| | Estimated RAT | $398.8\% \times 20.9\% = 83.3\%$ | $907.4\% \times 21.7\% = 196.7\%$ | $6.27\% \times 21.5\% = 1.35\%$ | $127.6\% \times 25.5\% = 32.54\%$ |
| | Measured RAT | 85.5% | 198.1% | 1.71% | 37.7% |

TABLE V
RESULTS OF REPLACING PRUNING CRITERIA

| | Pruning criteria | L7 Irish 3C | | | | L8 Biome 3C | | | | RIT-18 | | | |
|---|---|---|---|---|---|---|---|---|---|---|---|---|---|
| | | Index | mIoU % | CAP % | DCA % | Index | mIoU % | CAP % | DCA % | Index | mA % | CAP % | DCA % |
| OSTA | Validation accuracy | 56 | 62.49 | 100.00 | +1.42 | 9 | 68.38 | 100.00 | +1.69 | 19 | 66.53 | 100.00 | +1.21 |
| Test 1 | Train accuracy | 34 | 60.75 | 94.90 | +0.40 | 16 | 66.73 | 96.94 | +0.81 | 14 | 65.45 | 100.00 | +1.57 |
| Test 2 | PCA | 42 | 58.53 | 41.14 | +0.85 | 90 | 64.56 | 76.33 | +0.92 | 8 | 65.00 | 97.17 | +1.23 |
| Test 3 | Entropy | 49 | 58.23 | 31.92 | +0.77 | 65 | 65.72 | 89.11 | +0.37 | 20 | 64.01 | 66.00 | +1.89 |

| | | L7 Irish 2C | | | | L8 Biome 2C | | | | Semantic3D | | | |
|---|---|---|---|---|---|---|---|---|---|---|---|---|---|
| | | Index | mIoU % | CAP % | DCA % | Index | mIoU % | CAP % | DCA % | Index | mIoU % | CAP % | DCA % |
| OSTA | Validation accuracy | 50 | 75.40 | 100.00 | +1.69 | 23 | 87.63 | 100.00 | +0.42 | 36 | 70.86 | 100.00 | +1.08 |
| Test 1 | Train accuracy | 53 | 73.65 | 98.02 | +1.16 | 28 | 87.40 | 90.76 | +0.49 | 21 | 69.61 | 93.32 | +0.62 |
| Test 2 | PCA | 42 | 71.47 | 63.17 | +0.71 | 90 | 83.59 | 11.67 | +0.33 | 51 | 67.70 | 60.04 | +0.46 |
| Test 3 | Entropy | 49 | 71.98 | 73.75 | +0.39 | 65 | 85.45 | 35.42 | +0.75 | 1 | 63.12 | 12.69 | +1.25 |

The quantitative segmentation accuracies are summarised in Table III, together with the index of the SCC and the DCA where applicable. It was found that even for the same object (i.e., cloud), SGS selected different channel combinations for different granularity of annotation (L7 Irish 3C verse L7 Irish 2C and L8 Biome 3C verse L8 Biome 2C). This demonstrates the importance of task adaptive capacity for channel selection methods. In addition, it was noticed that the SCC were different for OSTA and SGS. This observation was further investigated in Section III.C.

## B. Efficiency of OSTA

The efficiency performances of OSTA are shown in Table IV. It was observed that the RAM of OSTA was equal to zero. This was because the calculations required for the proposed pruning method were the same as the forward calculations for training and did not require additional GPU memory.

The calculation for estimating RATs and the measured RATs are shown in Table IV. The number of equivalent pruning iterations required for L7 Irish, L8 Biome, RIT-18 and Semantic3D was calculated to be 398.8%, 196.7%, 6.27% and 127.6% of the number of total training iterations, respectively. The final estimated RAT values were much smaller than these values (i.e., 83.3%, 196.7%, 1.35% and 32.54%, respectively) because a pruning iteration took much less time than a training iteration. The estimated RATs were close to, but slightly lower than the actual measured values. This is because parallelism was not perfect during the actual calculation. Nevertheless, this proves that the time consumption of OSTA was predictable.

## C. Ablation study

### 1) Replacing for pruning stage

The pruning had two components, including the pruning criteria and the pruning strategy. The effect of using different pruning criteria on the final accuracy was first tested. The test results are shown in Table V. The highest accuracies were achieved when using the validation accuracy as the pruning criterion. In addition, these SCCs all achieved higher accuracies in Tests 1 to 3 than they did in the SGS (i.e., positive DCAs). This suggested that the pruning criteria were not the reason for obtaining a positive DCA.

TABLE VI
RESULTS OF REPLACING PRUNING STRATEGY

| | Pruning strategy | L7 Irish 3C | | | | L8 Biome 3C | | | | RIT-18 | | | |
|---|---|---|---|---|---|---|---|---|---|---|---|---|---|
| | | Index | mIoU % | CAP % | DCA % | Index | mIoU % | CAP % | DCA % | Index | mA % | CAP % | DCA % |
| OSTA | Discrete forward-only pruning strategy | 56 | 62.49 | 100.00 | +1.42 | 9 | 68.38 | 100.00 | +1.69 | 19 | 66.53 | 100.00 | +1.21 |
| Test 4 | Ranking at the end of stage 1 | 13 | 60.56 | 92.52 | +0.73 | 16 | 66.51 | 95.56 | +0.59 | 16 | 65.24 | 99.30 | +0.95 |
| | | L7 Irish 2C | | | | L8 Biome 2C | | | | Semantic3D | | | |
| | | Index | mIoU % | CAP % | DCA % | Index | mIoU % | CAP % | DCA % | Index | mIoU % | CAP % | DCA % |
| OSTA | Discrete forward-only pruning strategy | 50 | 75.40 | 100.00 | +1.69 | 23 | 87.63 | 100.00 | +0.42 | 36 | 70.86 | 100.00 | +1.08 |
| Test 4 | Ranking at the end of stage 1 | 39 | 73.95 | 98.57 | +1.37 | 32 | 87.38 | 100.00 | +0.38 | 11 | 69.95 | 96.16 | +0.75 |

TABLE VII
RESULTS OF REMOVING LINEAR WARMUP AND/OR SUPERNET TRAINING STAGE

| | Removing | | L7 Irish 3C | | | | L8 Biome 3C | | | | RIT-18 | | | |
|---|---|---|---|---|---|---|---|---|---|---|---|---|---|---|
| | Linear warmup | Supernet training | Index | mIoU % | CAP % | DCA % | Index | mIoU % | CAP % | DCA % | Index | mA % | CAP % | DCA % |
| OSTA | N | N | 56 | 62.49 | 100.00 | +1.42 | 9 | 68.38 | 100.00 | +1.69 | 19 | 66.53 | 100.00 | +1.21 |
| Test 5 | Y | N | 56 | 62.36 | 100.00 | +1.29 | 2 | 68.31 | 100.00 | +1.50 | 19 | 66.25 | 100.00 | +0.93 |
| Test 6 | N | Y | 53 | 61.06 | 98.13 | +0.63 | 23 | 67.49 | 98.90 | +0.51 | 19 | 65.77 | 100.00 | +0.45 |
| Test 7 | Y | Y | 26 | 60.57 | 92.63 | -0.02 | 1 | 66.49 | 95.49 | +0.28 | 10 | 64.69 | 92.00 | +0.04 |
| | | | L7 Irish 2C | | | | L8 Biome 2C | | | | Semantic3D | | | |
| | | | Index | mIoU % | CAP % | DCA % | Index | mIoU % | CAP % | DCA % | Index | mIoU % | CAP % | DCA % |
| OSTA | N | N | 50 | 75.40 | 100.00 | +1.69 | 23 | 87.63 | 100.00 | +0.42 | 36 | 70.86 | 100.00 | +1.08 |
| Test 5 | Y | N | 33 | 74.13 | 98.84 | +0.56 | 2 | 87.54 | 100.00 | +0.26 | 36 | 70.66 | 100.00 | +0.87 |
| Test 6 | N | Y | 33 | 73.70 | 98.09 | +0.13 | 81 | 87.41 | 100.00 | +0.22 | 56 | 70.59 | 100.00 | +0.32 |
| Test 7 | Y | Y | 12 | 73.44 | 94.54 | +0.04 | 96 | 87.34 | 100.00 | +0.14 | 55 | 70.11 | 97.29 | -0.14 |

TABLE VIII
RESULTS OF DIRECTLY FINE-TUNING THE TRAINED SUPERNET WITH SELECTED COMBINATIONS BY OSTA

| | Direct fine-tune the trained supernet on channel combination | L7 Irish 3C | | | | L8 Biome 3C | | | | RIT-18 | | | |
|---|---|---|---|---|---|---|---|---|---|---|---|---|---|
| | | Index | mIoU % | CAP % | DCA % | Index | mIoU % | CAP % | DCA % | Index | mA % | CAP % | DCA % |
| OSTA | - | 56 | 62.49 | 100.00 | +1.42 | 9 | 68.38 | 100.00 | +1.69 | 19 | 66.53 | 100.00 | +1.21 |
| Test 8 | Selected by OSTA | 56 | 62.06 | 100.00 | +1.09 | 9 | 68.13 | 100.00 | +1.44 | 19 | 66.05 | 100.00 | +0.73 |
| | | L7 Irish 2C | | | | L8 Biome 2C | | | | Semantic3D | | | |
| | | Index | mIoU % | CAP % | DCA % | Index | mIoU % | CAP % | DCA % | Index | mIoU % | CAP % | DCA % |
| OSTA | - | 50 | 75.40 | 100.00 | +1.69 | 23 | 87.63 | 100.00 | +0.42 | 36 | 70.86 | 100.00 | +1.08 |
| Test 8 | Selected by OSTA | 50 | 74.87 | 99.94 | +1.16 | 23 | 87.43 | 100.00 | +0.22 | 36 | 70.48 | 100.00 | +0.70 |

In the second ablation experiment, a very aggressive pruning strategy was tested. It ranked the validation accuracy for all channel combinations once at the end of the supernet training (i.e., stage 1) and the combination with the highest accuracy was selected for subsequent fine-tuning. The original iterations used for pruning were merged into the fine-tuning stage in this strategy. The results in Table VI suggested that an overaggressive pruning strategy could lead to a remarkable drop in the final accuracy. Nevertheless, relatively large positive DCAs were still obtained in this ablation experiment, suggesting that the main cause for the positive DCA would not be in the pruning stage.

2) **Replacing for supernet training stage**

The results of ablation on the supernet training stage are shown in Table VII. When the linear warmup was removed, the poly learning rate policy was used for all stages of OSTA. Meanwhile, when the supernet training was removed, OSTA was started directly from the pruning stage. From the results of Test 5-7, it is clear that the entire supernet training stage was the key to achieving the positive DCA.

Based on this finding, an additional set of tests was conducted. The selected combinations by OSTA were used to directly fine-tune the trained supernet. The results in Table VIII confirmed that using the trained supernet as a pre-trained model can improve DCA, and indicated that the pruning stage can further boost DCA.

3) **Replacing for network initialization methods**

As shown in Table VII, pruning from different initial conditions may select different channel combinations, which can be interpreted from the following perspective. For a given dataset and SSN, the "best" channel combination may not be fixed for different initial values of parameters used. Therefore, two additional network initialization methods were tested in this ablation study, including a SSN fine-tuning on the Cityscapes [56] (based on parameters trained on ImageNet) and a randomly initialized SSN. Due to a tremendous amount of work required for benchmarking, only the Semantic3D data was tested. Semantic3D is chosen because it contains scenes similar to that in the Cityscapes. It is of interest to test the effect of using a dataset from similar scenes to pre-train the SSN on segmentation accuracy.

TABLE IX
RESULTS OF REPLACING NETWORK INITIALIZATION METHODS

| | | ImageNet | | | | Cityscapes | | | | Random | | | |
|---|---|---|---|---|---|---|---|---|---|---|---|---|---|
| | | Index | mIoU % | CAP % | DCA % | Index | mIoU % | CAP % | DCA % | Index | mIoU % | CAP % | DCA % |
| DF | | - | 61.44 | 4.47 | - | - | 64.58 | 17.86 | - | - | 45.77 | 86.31 | - |
| Feature extraction | PCA | - | 66.13 | 49.68 | - | - | 63.85 | 14.17 | - | - | 35.46 | 9.69 | - |
| | LC-Net | - | 69.67 | 93.79 | - | - | 70.30 | 96.76 | - | - | 46.79 | 91.84 | - |
| Channel selection | BS-Nets | 34 | 68.72 | 80.36 | 0 | 34 | 69.63 | 87.50 | 0 | 34 | 38.33 | 33.93 | 0 |
| | ONR | 27 | 62.96 | 10.71 | 0 | 27 | 64.08 | 16.07 | 0 | 27 | 33.08 | 7.14 | 0 |
| | DARecNet-BS | 43 | 65.26 | 35.71 | 0 | 43 | 67.41 | 53.57 | 0 | 43 | 45.79 | 87.50 | 0 |
| | Top 10 in SGS | 56 | 70.27 | 100.00 | 0 | 36 | 71.19 | 100.00 | 0 | 26 | 52.70 | 100.00 | 0 |
| | | 55 | 70.25 | 98.21 | 0 | 46 | 70.88 | 98.21 | 0 | 46 | 50.08 | 98.21 | 0 |
| | | 46 | 69.98 | 96.43 | 0 | 55 | 70.17 | 96.43 | 0 | 11 | 49.62 | 96.43 | 0 |
| | | 36 | 69.78 | 94.64 | 0 | 56 | 70.13 | 94.64 | 0 | 36 | 47.13 | 94.64 | 0 |
| | | 52 | 69.55 | 92.86 | 0 | 21 | 70.02 | 92.86 | 0 | 25 | 46.83 | 92.86 | 0 |
| | | 25 | 69.30 | 91.07 | 0 | 19 | 69.86 | 91.07 | 0 | 6 | 46.76 | 91.07 | 0 |
| | | 11 | 69.20 | 89.29 | 0 | 14 | 69.82 | 89.29 | 0 | 33 | 46.56 | 89.29 | 0 |
| | | 21 | 68.99 | 87.50 | 0 | 34 | 69.63 | 87.50 | 0 | 43 | 45.79 | 87.50 | 0 |
| | | 19 | 68.80 | 85.71 | 0 | 5 | 69.57 | 85.71 | 0 | 56 | 45.76 | 85.71 | 0 |
| | | 44 | 68.76 | 83.93 | 0 | 39 | 69.46 | 83.93 | 0 | 21 | 45.34 | 83.93 | 0 |
| | OSTA | 36 | 70.86 | 100.00 | +1.08 | 46 | 71.39 | 100.00 | +0.51 | 26 | 55.97 | 100.00 | +3.27 |

TABLE X
SUMMARY OF RECURRING CHANNEL COMBINATIONS IN THE TOP 10 OF SGS IN TABLE IX

| Index | Channels | | | CAP % | | | | |
|---|---|---|---|---|---|---|---|---|
| | | | | ImageNet | Cityscapes | Random | Average | Standard deviation |
| 11 | R | B | De | 89.29 | 67.86 | 96.43 | 84.52 | 14.87 |
| 36 | G | Ze | De | 94.64 | 100.00 | 94.64 | 96.43 | 3.09 |
| 46 | B | Ze | De | 96.43 | 98.21 | 98.21 | 97.62 | 1.03 |
| 55 | Z | Ze | De | 98.21 | 96.43 | 75.00 | 89.88 | 12.92 |
| 56 | D | Ze | De | 100.00 | 94.64 | 85.71 | 93.45 | 7.22 |

TABLE XI
DETAILED SEGMENTATION RESULTS FOR COMBINATION 11, 36, 46, 55 AND 56 ON SEMANTIC3D, WITH DIFFERENT PRETRAINED WEIGHTS

| Pretrained weights | Index | CAP % | mIoU % | IoU % | | | | | | | |
|---|---|---|---|---|---|---|---|---|---|---|---|
| | | | | Man-made terrain | Natural terrain | High vegetation | Low vegetation | Buildings | Hard scape | Scanning artefacts | Cars |
| ImageNet | 11 | 89.29 | 69.20 | 91.81 | 83.67 | 88.17 | 62.04 | 89.27 | 31.47 | 36.17 | 71.01 |
| | 36 | 94.64 | 69.78 | 92.49 | 83.91 | 88.84 | 63.85 | 88.00 | 30.57 | 39.21 | 71.16 |
| | 46 | 96.43 | 69.98 | 92.83 | 83.60 | 87.80 | 66.66 | 88.40 | 35.08 | 35.06 | 70.43 |
| | 55 | 98.21 | 70.25 | 92.78 | 84.83 | 89.38 | 65.61 | 88.57 | 31.00 | 39.22 | 70.64 |
| | 56 | 100.00 | 70.27 | 92.66 | 86.71 | 90.19 | 70.23 | 88.00 | 34.64 | 24.16 | 75.23 |
| Cityscapes | 11 | 67.86 | 68.70 | 90.07 | 81.15 | 86.29 | 60.17 | 87.41 | 31.69 | 40.73 | 72.08 |
| | 36 | 100.00 | 71.19 | 92.43 | 83.66 | 88.94 | 62.38 | 89.93 | 36.05 | 43.35 | 72.78 |
| | 46 | 98.21 | 70.88 | 92.42 | 83.86 | 88.46 | 61.47 | 89.96 | 36.42 | 43.01 | 71.46 |
| | 55 | 96.43 | 70.17 | 92.97 | 83.46 | 87.87 | 62.56 | 87.60 | 34.49 | 39.17 | 73.21 |
| | 56 | 94.64 | 70.13 | 92.31 | 85.42 | 89.45 | 61.93 | 90.20 | 36.26 | 33.69 | 71.75 |
| Random | 11 | 96.43 | 49.62 | 82.28 | 47.90 | 84.62 | 33.08 | 77.34 | 14.67 | 28.70 | 28.39 |
| | 36 | 94.64 | 47.13 | 62.99 | 33.83 | 85.35 | 30.75 | 76.38 | 20.30 | 31.6 | 35.88 |
| | 46 | 98.21 | 50.08 | 76.69 | 51.08 | 82.90 | 31.51 | 78.51 | 21.12 | 27.46 | 31.36 |
| | 55 | 75.00 | 44.89 | 77.68 | 33.78 | 76.35 | 23.78 | 72.02 | 20.27 | 27.87 | 27.35 |
| | 56 | 85.71 | 45.76 | 81.61 | 44.43 | 69.63 | 26.98 | 75.74 | 20.34 | 24.71 | 22.61 |

The results are summarised in Table IX, which supported the speculation that using different pre-training parameters would lead to changes in the "best" channel combination. Nevertheless, OSTA achieved the highest accuracy for all the initialisation cases tested, again proving its task adaptive capability. In addition, almost all methods achieved better absolute segmentation accuracies (mIoU) using the cityscapes pre-trained SSN than using the ImageNet pre-trained SSN. This illustrates the importance of similarity between the fine-tuning and pre-training data for the fine-tuning accuracy.

IV. DISCUSSION

A. Most robust combination

It was found that some channel combinations from SGS in

TABLE XII

REPRODUCTION OF TABLE XI, RE-ARRANGED IN THE VERTICAL DIRECTION ACCORDING TO THE COMBINATION USED. THE BOLDED ONE INDICATE THE BEST ACCURACY FOR THAT CLASS WERE ACHIEVED BY USING CORRESPONDING COMBINATION

| Index | Pretrained weights | CAP % | mIoU % | IoU % | | | | | | | |
|---|---|---|---|---|---|---|---|---|---|---|---|
| | | | | Man-made terrain | Natural terrain | High vegetation | Low vegetation | Buildings | Hard scape | Scanning artefacts | Cars |
| 11 | ImageNet | 89.29 | 69.20 | 91.81 | 83.67 | 88.17 | 62.04 | **89.27** | 31.47 | 36.17 | 71.01 |
| | Cityscapes | 67.86 | 68.70 | 90.07 | 81.15 | 86.29 | 60.17 | 87.41 | 31.69 | 40.73 | 72.08 |
| | Random | 96.43 | 49.62 | **82.28** | 47.90 | 84.62 | **33.08** | 77.34 | 14.67 | 28.70 | 28.39 |
| 36 | ImageNet | 94.64 | 69.78 | 92.49 | 83.91 | 88.84 | 63.85 | 88.00 | 30.57 | **39.21** | 71.16 |
| | Cityscapes | 100.00 | **71.19** | 92.43 | 83.66 | 88.94 | 62.38 | 89.93 | 36.05 | **43.35** | 72.78 |
| | Random | 94.64 | 47.13 | 62.99 | 33.83 | **85.35** | 30.75 | 76.38 | 20.30 | 31.60 | **35.88** |
| 46 | ImageNet | 96.43 | 69.98 | **92.83** | 83.60 | 87.80 | 66.66 | 88.40 | **35.08** | 35.06 | 70.43 |
| | Cityscapes | 98.21 | 70.88 | 92.42 | 83.86 | 88.46 | 61.47 | 89.96 | **36.42** | 43.01 | 71.46 |
| | Random | 98.21 | **50.08** | 76.69 | **51.08** | 82.90 | 31.51 | **78.51** | **21.12** | 27.46 | 31.36 |
| 55 | ImageNet | 98.21 | 70.25 | 92.78 | 84.83 | 89.38 | 65.61 | 88.57 | 31.00 | 39.22 | 70.64 |
| | Cityscapes | 96.43 | 70.17 | **92.97** | 83.46 | 87.87 | **62.56** | 87.60 | 34.49 | 39.17 | **73.21** |
| | Random | 75.00 | 44.89 | 77.68 | 33.78 | 76.35 | 23.78 | 72.02 | 20.27 | 27.87 | 27.35 |
| 56 | ImageNet | 100.00 | **70.27** | 92.66 | **86.71** | 90.19 | 70.23 | 88.00 | 34.64 | 24.16 | **75.23** |
| | Cityscapes | 94.64 | 70.13 | 92.31 | **85.42** | 89.45 | 61.93 | **90.20** | 36.26 | 33.69 | 71.75 |
| | Random | 85.71 | 45.76 | 81.61 | 44.43 | 69.63 | 26.98 | 75.74 | 20.34 | 24.71 | 22.61 |

TABLE XIII

TOP 10 IN SGS FOR CLOUD DETECTION BENCHMARK DATA

| | L7 Irish 3C | | | | L7 Irish 2C | | | | L8 Biome 3C | | | | L8 Biome 2C | | | |
|---|---|---|---|---|---|---|---|---|---|---|---|---|---|---|---|---|
| | Index | Band 1 | Band 2 | Band 3 | Index | Band 1 | Band 2 | Band 3 | Index | Band 1 | Band 2 | Band 3 | Index | Band 1 | Band 2 | Band 3 |
| Top 10 in SGS | 50 | NIR | TLG | THG | 56 | TLG | THG | MIR | 3 | CA | B | NIR | 106 | NIR | SWIR2 | T1 |
| | 56 | TLG | THG | MIR | 50 | NIR | TLG | THG | 10 | CA | G | NIR | 22 | CA | NIR | SWIR1 |
| | 5 | B | G | THG | 33 | G | SWIR | MIR | 23 | CA | NIR | SWIR2 | 107 | NIR | SWIR2 | T2 |
| | 25 | G | R | THG | 12 | B | NIR | SWIR | 2 | CA | B | R | 23 | CA | NIR | SWIR2 |
| | 26 | G | R | MIR | 27 | G | NIR | SWIR | 9 | CA | G | R | 96 | R | SWIR2 | T1 |
| | 53 | SWIR | TLG | THG | 39 | R | NIR | THG | 22 | CA | NIR | SWIR1 | 81 | G | SWIR2 | T1 |
| | 34 | G | TLG | THG | 26 | G | R | MIR | 74 | G | NIR | T1 | 103 | NIR | SWIR1 | T1 |
| | 6 | B | G | MIR | 38 | R | NIR | TLG | 1 | CA | B | G | 60 | B | SWIR2 | T1 |
| | 15 | B | NIR | MIR | 53 | SWIR | TLG | THG | 37 | B | G | R | 27 | CA | SWIR1 | SWIR2 |
| | 39 | R | NIR | THG | 34 | G | TLG | THG | 16 | CA | R | NIR | 57 | B | SWIR1 | T1 |

Table IX achieved promising CAPs regardless of the initialisation values used for the parameters, which are summarised in Table X. Among these combinations, the combination 46 achieved excellent CAPs for all cases with an average value of 97.62%. The channel combination with such a characteristic is referred to as the most robust combination (MRC). Designing selection criteria for MRC is meaningful future work. The detailed segmentation results for the channel combinations in Table X are shown in Table XI, which might provide some inspiration for readers. For example, by changing the dimension used for sorting (Table XII), it was seen that combinations 46 and 36 dominated the highest segmentation accuracy for the two most difficult classes, i.e., hard scape and scanning artefacts, respectively.

## B. Coastal aerosol band for cloud detection

The CA band has always been ignored in the field of cloud detection. When L8 Biome was established, the CA band was not used to label clouds, as stated in [39] that "Band 1 (coastal aerosol) was never used". However, our study suggested that CA might be an important channel for cloud detection. As shown in Table XIII, the CA band was frequently present in the top 10 of all SGS channel combinations for the L8 Biome benchmark data. In particular, 8 of those top 10 combinations included the CA band for segmenting clouds into thin and thick clouds. In the future, it would be of interest to use methods such as Bradley-Terry model [57] to statistically analyse the error matrices generated by different channel combinations. New index might be developed for cloud detection. In addition, we conjecture that for other remote sensing tasks, "CA band" could also exist.

## C. Training with channel combinations other than the selected one

It was found that both the supernet training and the pruning stages boosted DCA. These two stages shared common operations in that they both used channel combinations other than SCC to train the SSN. Training in this way could force the SSN to learn to use channel-invariant features for semantic segmentation. This is the reason why SCC in OSTA can achieve higher accuracy than training SSN with SCC alone (i.e., in SGS). This may provide insight for the development of new methods for model pre-training and/or data augmentation. In addition, we conjecture that this mechanism can be used to reduce the domain-shift problem caused by the different spectrums used in image sensors. Integration of this mechanism with existing methods [58]–[60] may lead to better solutions for domain adaptation.

*D. Limitation of OSTA*

The major limitation of OSTA is that it still requires the prerequisite setting of dimension reduction target. Therefore, developing a channel selection method that can automatically determine the optimal number of channels is meaningful future work.

A complementary future work is to develop pre-trained models based on multichannel image datasets. This will not only benefit the task of channel selection, but will greatly facilitate the advancement of the field of multichannel image processing.

## V. CONCLUSION

It is always desirable to use fewer feature channels to achieve higher semantic segmentation accuracy. Although many channel selection methods have been developed to achieve this aim, they have several limitations. Limitations affecting the semantic segmentation accuracy come from the use of selection criteria other than segmentation accuracy and the use of evaluation of individual channels to select channel combinations. Meanwhile, the limitation affecting efficiency comes from the repetitive training of classifier(s). A one-shot task-adaptive (OSTA) channel selection method was proposed in this study to provide a potential solution to these limitations. OSTA is based on the concept of pruning from a supernet, which integrates the channel selection and SSN training processes, thus avoiding repetitive training of SSN. The limitations affecting accuracy are addressed by the use of the semantic segmentation accuracy of different channel combinations on the validation set as the pruning criterion in OSTA. The effectiveness and efficiency of OSTA were tested using four datasets, including L7 Irish, L8 Biome, RIT-18 and Semantic3D. OSTA achieved the highest semantic segmentation accuracies in all benchmark tests. Compared to a single training session of SSN, OSTA did not require extra memory footprint, and took a minimum of 1.71% to a maximum of 198.1% extra time (predictable) to select the best 3-channel combination for four datasets tested.

To the best of our knowledge, OSTA was the first channel selection method that produced a semantic segmentation accuracy exceeding the highest accuracy obtained by exhaustive tests of channel combinations. Our experiments suggested that this was because training the SSN with extra channel combinations could improve the semantic segmentation accuracy. This mechanism can potentially be used to develop new pre-training/data augmentation methods.

Our experiments also revealed that in addition to the "best channel combinations", there was a most robust channel combination that achieved excellent accuracy performance regardless of the network parameter initialisation method used. It is recommended that future work could be devoted to design selection criteria for this type of channel combination.

It was also interesting to find out that the coastal aerosol band was important for cloud detection. We recommend to investigate on new cloud detection methods in the future based on this finding.

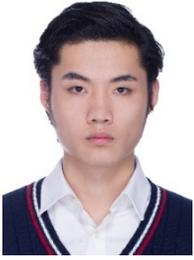

**Yuanzhi Cai** (Member, IEEE) received the B.E. degree in civil engineering from the University of Liverpool, UK, in 2019.

He is currently pursuing the Ph.D. degree at the same University. His research interests include the classification and segmentation of remote sensing data.

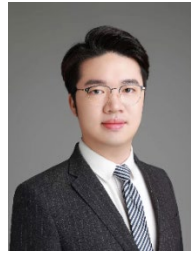

**Hong Huang** received his B.E. degree in civil engineering from the University of Liverpool in 2018.

He is currently working on his Ph.D. program at the same university. His research areas are mainly focused on 3D reconstruction, infrared imaging, semantic segmentation of point clouds, and building envelope structural monitoring.

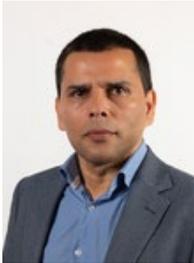

**Jagannath Aryal** (Member, IEEE) received the Ph.D. degree in optimization and systems modeling from Centre for Advanced Computational Solutions (C-fACS), Lincoln University, Lincoln, New Zealand, in 2010.

He is currently an Associate Professor in Digital Infrastructure Engineering with the Faculty of Engineering and Information Technology (FEIT), Department of Infrastructure Engineering, University of Melbourne, Melbourne, Australia. He leads the Remote Sensing and Photogrammetry Commission of SSSI, Australia as a national chair. His research interests and contributions include advancing the knowledge in Earth Observation Science, Digital Information Transformation, and Disaster Management using intelligent modelling approaches such as deep learning, geographic object-based artificial intelligence (GEOAI), resilience engineering, and spatial statistics. The application areas include urban systems, and urban features and disaster risk reductions.

Dr Aryal serves in IEEE Transactions on Geoscience and Remote Sensing as an associate editor. He also serves in the editorial board of Journal of Spatial Science of Taylor and Francis Group. He is a reviewer of remote sensing-related journals, including the IEEE and Elsevier Science Journals.

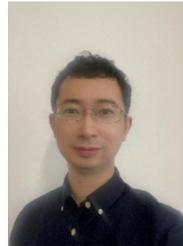

**Lei Fan** (Member, IEEE) received the Ph.D. degree from the University of Southampton, Southampton, UK, in 2018.

He is currently an Assistant Professor within Department of Civil Engineering at Xi'an Jiaotong Liverpool University, Suzhou, China. His main research interests include lidar and photogrammetry techniques, point cloud, machine learning, deformation monitoring, semantic segmentations, monitoring of civil engineering structures and geohazards.

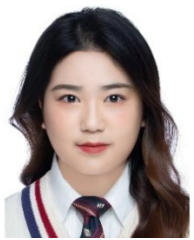

**Yuan Fang** received the B.E. degree in civil engineering from the University of Liverpool, UK, in 2020.

She is now working toward her PhD degree at the same University. Her research interests include deep learning and data fusion of satellite images.